\pdfoutput=1

\documentclass[11pt]{article}

\usepackage{acl}

\usepackage{graphicx}
\usepackage{times}
\usepackage{latexsym}
\usepackage{multirow}
\usepackage{wrapfig}
\usepackage{amsmath}
\usepackage{tabto}
\usepackage{bm}
\usepackage[linguistics]{forest}
\usepackage{booktabs}

\def\red{\color{red}}
\def\blue{\color{blue}}
\newcommand{\temv}[1]{\mbox{\em{\Huge \textbf{#1}}}}

\usepackage[T1]{fontenc}

\usepackage[utf8]{inputenc}

\usepackage{microtype}

\title{Compositional Task-Oriented Parsing as Abstractive Question Answering}

\author{Wenting Zhao$^{1}$ \thanks{ \hspace{2pt} Work done during an internship at Amazon Alexa AI.} \and Konstantine Arkoudas$^{2}$ \and Weiqi Sun$^{2}$ \and Claire Cardie$^{1}$\\
  $^1$Cornell University\hspace{14pt}$^2$Amazon Alexa AI \\
  \texttt{\{wzhao,cardie\}@cs.cornell.edu}, \texttt{\{arkoudk,weiqisun\}@amazon.com}
  }

\begin{document}
\maketitle
\begin{abstract}
Task-oriented parsing (TOP) aims to convert natural language into machine-readable representations of specific tasks, such as setting an alarm. A popular approach to TOP is to apply seq2seq models to generate linearized parse trees. A more recent line of work argues that pretrained seq2seq models are better at generating outputs that are themselves natural language, so they replace linearized parse trees with canonical natural-language paraphrases that can then be easily translated into parse trees, resulting in so-called naturalized parsers. In this work we continue to explore naturalized semantic parsing by presenting a general reduction of TOP to abstractive question answering that overcomes some limitations of canonical paraphrasing. Experimental results show that our QA-based technique outperforms state-of-the-art methods in full-data settings while achieving dramatic improvements in few-shot settings.
\end{abstract}

\section{Introduction}
Task-oriented parsing (TOP) takes an utterance as input and generates an unambiguous specification of a task that can be executed by machine~\cite{gupta2018semantic}.
Traditional approaches to TOP treat the task as an instance of slot filling~\cite{liu16c_interspeech}, first classifying the intent of the utterance as a whole and then tagging tokens with slot labels that identify relevant entities (such as numbers, persons, dates or times, organizations, and so on).
However, this approach only works for simple utterances that have flat rather than compositional semantics.
That is, slot-filling approaches cannot produce nested (or ``hierarchical'') meaning representations, such as the one shown in Fig~\ref{fig:navigation_example}, where slots and intents can be composite and contain other slots (or intents) as proper parts. 


A straightforward way to handle compositional semantics is to formulate TOP as a seq2seq problem, where the input sequence is the utterance and the output sequence is a linearized representation of a semantic tree (shown in the bottom part of Fig~\ref{fig:navigation_example}).
The recent development of high-performing transformer-based pretrained language models (PLMs)~\cite{lewis2020bart,raffel2020exploring,NEURIPS2020_1457c0d6} that can be fine-tuned on specific tasks (such as a particular TOP dataset) has made this formulation feasible.

This approach has achieved state-of-the-art performance on a variety of TOP datasets~\cite{rongali2020don,aghajanyan2020conversational, chen2020low}.
However, the output sequences are typically not expressed in natural language but rather in a mixture of natural language and special symbols (such as {\tt [SL:}) that were not seen during pretraining and have no meaning to the PLMs.
Depending on the task, the output sequences may hardly contain any natural language at all (for example, if the outputs are SQL queries).
It would seem reasonable to conjecture that performance would improve if we could reformulate TOP as a more conventional NLP task in which both the inputs and the outputs are expressed in natural language, as such a reformulation might be better able to leverage the PLM's pretraining. 


\begin{figure}[t!]
  \small \textbf{\textit{Utterance:}} Look up directions to the nearest parking near S Beritania Street.\\
  \hspace{.5cm}
  \small \textbf{\textit{Semantic parse tree of the utterance:}}
  \vspace*{-.4cm}
  \begin{center}
    \begin{tabular}{c}
    \resizebox{.455\textwidth}{!}{
        {\large \begin{forest}for tree={font=\LARGE}
          [IN:GET\textunderscore DIRECTIONS [SL:DESTINATION [IN:GET\textunderscore LOCATION [SL:LOCATION\textunderscore MODIFIER [\temv{nearest}]] [SL:CATEGORY\textunderscore LOCATION [\temv{parking}]] [SL:LOCATION\textunderscore MODIFIER [IN:GET\textunderscore LOCATION [SL:SEARCH\textunderscore RADIUS [\temv{near}]] [SL:LOCATION [\temv{S Beritania Street}]]]]]]]
        \end{forest}
        }}
    \end{tabular}
\end{center}
    \parbox{0.46\textwidth}{%
    \small
      \textbf{\textit{Linarized parse tree:}}\vspace*{.05cm}\\
      {[IN:GET\textunderscore DIRECTIONS \\
      \hspace*{.3cm} {[SL:DESTINATION \\
      \hspace*{.6cm} {[IN:GET\textunderscore LOCATION \\
      \hspace*{.9cm} {[SL:LOCATION\textunderscore MODIFIER nearest]} \\ 
      \hspace*{.9cm} {[SL:CATEGORY\textunderscore LOCATION parking]} \\ 
      \hspace*{.9cm} {[SL:LOCATION\textunderscore MODIFIER \\
      \hspace*{1.2cm} {[IN:GET\textunderscore LOCATION \\
      \hspace*{1.5cm} {[SL:SEARCH\textunderscore RADIUS near]}\\
      \hspace*{1.5cm} {[SL:LOCATION S Beritania Street]} ]} ]} ]} ]} ]}
    }
\caption{A sample utterance and its semantic parse tree in the Topv2 dataset, where nodes starting with ``IN:'' are intents and nodes starting with ``SL:'' are slots. A parse tree always has an intent node at the root. An intent typically corresponds to a verb and can be viewed as an action, with a sequence of slots as its arguments. A slot may have additional intents nested in it, recursively. A linearized parse tree that can be processed by seq2seq models is shown at the bottom.}
\label{fig:navigation_example}
\end{figure}

To bridge this gap, \citet{semanticmachines-2021-emnlp} reduced TOP to a canonical-paraphrasing task.
They first fine-tune a PLM to map a natural utterance $u$ to another {\em canonical utterance\/} $u'$, where canonical utterances belong to a controlled fragment of the relevant natural language.
Then $u'$ can be translated into the desired meaning representation (semantic tree) via a context-free grammar that can parse all and only canonical utterances.
However, this approach has a major limitation: canonical fragments can be defined and parsed by hand-written grammars only in closed-world domains where the set of underlying entities is fixed and known in advance (e.g., in a domain where people ask questions about basketball players, all of whom are statically known).
That condition rarely holds in task-oriented domains, and therefore this method is not applicable to datasets like Topv2.
Moreover, as input utterances grow more structurally complex, the associated canonical utterances become much longer, and generating long sequences is known to be challenging for PLMs~\cite{guo2018long}.

In this paper we focus on scenarios where it is not viable to specify a canonical grammar with a fixed set of rules.
We instead propose to formulate TOP as abstractive question answering, in such a way that answering all questions for a given utterance allows us to reconstruct the parse tree of that utterance. Specifically, we introduce \emph{single-turn QA} (ST QA for short), which poses one single query for a given utterance, and \emph{multi-turn QA} (MT QA), which dynamically constructs multiple questions for each utterance, depending on previous answers. Because single-turn QA asks only one question, it has lower latency; however, the model must generate a longer text representing the entire parse tree. By contrast, multi-turn QA generates shorter answers that are more straightforward to parse; however, all questions need to be answered correctly, and if there are dependencies between questions, they can't be run within the same batch. We study these two approaches and their tradeoffs in both full-data and low-resource settings.

To summarize our contributions: \vspace*{-.25cm}

\begin{enumerate}
\item We propose a general reduction of compositional TOP to abstractive QA and introduce two specific variants: single-turn QA and multi-turn QA, each with unique benefits. \vspace*{-.3cm}
\item We train the abstractive QA models with a masked span prediction (MSP) objective, one of the pretraining objectives of the seq2seq model, which is shown to yield very substantial improvements in few-shot scenarios. \vspace*{-.3cm}
\item We evaluate ST QA and MT QA on two public datasets, Topv2~\cite{chen2020low} and Pizza~\cite{pizzaDataset}, and show that our results improve on the state of the art by 3\% on full-data Topv2, 28\% for few-shot Topv2, and 7\% on few-shot Pizza.~\footnote{The QA datasets converted from the TOP format are available at \url{https://github.com/amazon-research/semantic-parsing-as-abstractive-qa}.}
\end{enumerate}

\section{Related Work}
Task-oriented parsing has been extensively studied in the literature. The most prevalent approach is seq2seq modeling, which maps utterances to their meaning representations,
typically expressed as a mixture of natural language and tokens such as brackets and artificial intent and slot identifiers~\cite{rongali2020don,zhou2021amr,aghajanyan2020conversational,shrivastava2021span,mansimov2021semantic}; we take this approach as  our main baseline.

Few-shot semantic parsing has also attracted wide interest.
\citet{chen2020low} applied a different training paradigm; they assumed there are several source domains with labeled data and adopted a first-order meta-learning algorithm, Reptile, to train their model. 

\citet{semanticmachines-2021-emnlp} argued that PLMs are better suited for directly generating natural language rather than task-specific meaning representations, and thus they fine-tune PLMs to generate canonical paraphrases, which can then be parsed by a context-free grammar to produce the corresponding semantic trees.  They further improve performance by  augmenting input sequences with similar examples  as prompts. \citet{RongaliLittleData2022} push that direction further by leveraging small amounts of unannotated data. 
We use canonical paraphrasing as one of our baselines. As we already noted in the introduction, canonical paraphrasing is not widely applicable in open-world task-oriented parsing; our QA-based approach overcomes that limitation. \citet{desai2021low} applied modifications to the linearized trees to make them more natural.

Inspired by the recent success of QA-driven approaches to a wide range of NLP tasks, such as dialogue state tracking~\cite{gao2019dialog}, named entity recognition~\cite{li2020unified}, and multi-task learning~\cite{mccann2018natural,du2021qa}, \citet{namazifar2021language} framed semantic parsing as an extractive QA task. This limits its scope to non-compositional semantic structures. Our work is the first to use QA for semantic parsing of arbitrarily nested and complex meaning representations. 
Moreover, in contrast to previous approaches, our formulation usually results in fewer questions.

\section{Reducing TOP to Abstractive QA}
\label{sec:method}
We now present a general method for reducing compositional TOP to abstractive QA.  Given an utterance, our goal is to recover its semantic parse tree by asking questions and parsing the answers returned by a QA model.
For this to succeed, all questions associated with an input utterance need to be answered correctly.
At a high level, we formulate questions so as to build the required parse tree in a top-down, left-to-right fashion: We first ask a question to determine the root node, and then we recursively proceed towards the leaves. We start by describing multi-turn QA; the single-turn case is discussed in Section~\ref{Sec:SingleTurnQA}.

Each QA instance is a triple consisting of a context, a question, and an answer.
We use the parse tree in Fig.~\ref{fig:navigation_example} to illustrate the corresponding (multi-turn) sequence of QA triples shown in Fig.~\ref{fig:qa_example_topv2}.
The context is the utterance provided by the user plus general information about the domain and/or state from previous turns; the question corresponds to a particular node of the parse tree; and the answer provides the content of that node.  We first extract the top-level intent ({\em get directions}), then the corresponding slots ({\em destination}), then the value of each slot ({\em destination is the nearest parking near S Beritania Street}), and then we start to recursively repeat this process on the phrase {\em the nearest parking near S Beritania Street} (by asking what is the intent included in that utterance segment). We note that our system is able to handle negative answers: If there isn't a nested intent in an utterance segment, the system returns \textit{none}.

\begin{figure}[t!]
\textbf{\textit{{\small Multi-turn QA:}}}\\
\noindent\fbox{%
    \parbox{0.46\textwidth}{%
        \small
        \textbf{Q}: \textbf{{\blue A user may intend to get directions, get distance, get estimated arrival time, get estimated departure time, get estimated duration, get road condition information, get route's information, get traffic information, get location, make unsupported navigation, or update directions.}} A user said, ``Look up directions to the nearest parking near S Beritania Street.'' What did the user intend to do?\\
        \textbf{A}: get directions\\
        \textbf{Q}: \textbf{{\blue The slots for get directions may be locations, arrival datetimes, road conditions to avoid, waypoints, amounts, paths, sources, travel methods, road conditions, waypoints to avoid, departure datetimes, paths to avoid, obstructions to avoid, and destinations}}. A user said ``Look up directions to the nearest parking near S Beritania Street.'' \textit{{\red The user's intent is to get directions.}} What are the slots?\\
        \textbf{A}: destination\\
        \textbf{Q}: A user said ``Look up directions to the nearest parking near S Beritania Street.'' \textit{{\red The user's intent is to get directions, and the slot is destination.}} What is the destination?\\
        \textbf{A}: the nearest parking near S Beritania Street\\
        \textbf{Q}: \textbf{{\blue The nested intent in destination may be get school's location, get home's location, get location, get event, and get workplace's location.}} A user said ``Look up directions to the nearest parking near S Beritania Street.'' \textit{{\red The user's intent is to get directions, and the destination is the nearest parking near S Beritania Street.}} Is there an intent included in ``the nearest parking near S Beritania Street?''\\
        \textbf{A}: get location\\
        {[$\cdots$]}
        }
    }
\textbf{\textit{{\small Single-turn QA:}}}\\
\noindent\fbox{%
    \parbox{0.46\textwidth}{%
        \small
        \textbf{Q}: \textbf{{\blue All possible intents from a user are [...], and slots could be [...].}} A user said, ``Look up directions to the nearest parking near S Beritania Street.'' What did the user intend to do?\\
        \textbf{A}: The user intended to get directions, where destination is nearest parking near S Beritania Street. The intent for ``nearest parking near S Beritania Street'' is to get location, where location's category is parking and location modifiers are near S Beritania Street; nearest. The intent for ``near S Beritania Street'' is get location, where location is S Beritania Street and search radius is near.
        }
    }
    \caption{Examples of multi-turn QA (top) and single-turn QA (bottom) for the utterance \textit{Look up directions to the nearest parking near S Beritania Street}. 
}
    \label{fig:qa_example_topv2}
\end{figure}

\subsection{State Tracking}
\label{sec:state_tracking}
The triples we have described are processed independently from one another. To provide richer information to the QA model, we include previous answers as additional context (shown in italicized red font in Fig.~\ref{fig:qa_example_topv2}). We represent a previous question-answer pair by combining them into a declarative sentence, and we stack all previous states together. This essentially encodes all parse-tree ancestors in natural language, which can potentially help the QA system resolve ambiguities.
For example, the nested intent in \emph{the nearest parking near S Beritania Street} could be different depending on whether this was a source or a destination.

\subsection{Incorporating Domain Metadata}
\label{sec:ontology}
A given domain has predefined semantics, i.e., a fixed number of intents, each of which has an associated set of slots that describe important aspects of user requests.
We incorporate metadata relevant to the current node as additional context (shown in blue bold font in Fig.~\ref{fig:qa_example_topv2}).
This effectively reduces the space of possible answers. For instance, while a domain may have a large number of intents, the intents that can appear at a particular node (e.g., at the root level) may be substantially fewer. And because we explicitly list all possible intents, the QA system can simply copy and paste the appropriate tokens, which is easier than searching over the entire vocabulary.

\subsection{Single-turn QA vs Multi-turn QA}
\label{Sec:SingleTurnQA}
We note that there is no need to limit ourselves to  one question for every node.
For the example of Fig.~\ref{fig:qa_example_topv2}, for instance, we don't ask one question for each child node of IN:GET\textunderscore LOCATION individually (e.g., what is the first/second/third slot); instead, we simply ask ``what are the slots?'' and the answer should be ``location modifier, location category.''
Another example is when we ask ``what is the location modifier?'', the answer being ``nearest; near S Beritania Street.''

On the extreme side, we could ask one question that would return an answer representing the entire parse tree, and this becomes somewhat similar to canonical paraphrasing~\cite{semanticmachines-2021-emnlp}. However, canonical paraphrasing assumes there is concrete grammar that specifies a controlled fragment of natural language (all and only the canonical utterances), which can be used to map sentences from that fragment into parse trees. That assumption often fails in open-world TOP domains; for example, when someone asks for directions, the destination could be expressed by an unbounded number of phrases (my parent's house, a restaurant that satisfies an arbitrary set of constraints, etc.) that cannot be specified a priori by a closed-form grammar. 

In the single-turn part of Fig.~\ref{fig:qa_example_topv2}, we show how we compress an entire parse tree into a single QA triple. The bold blue context is again encoding the domain's metadata. From left to right, the answer explicitly lists the relevant intents, their associated slots, and the tokens corresponding to the slots in a top-down direction. We deal with nested intents by recursively adding new sentences, which start with the tokens under which the intent is nested. 


While our approach is  more general than canonical paraphrasing, we still prefer canonicalization when possible, as the corresponding fragments tend to be more easily learnable. In the case of Pizza, for instance, a canonical grammar can be defined fairly straightforwardly.  We illustrate the use of such canonical utterances in combination with our QA approach in Fig.~\ref{fig:pizza_qas}.
For multi-turn QA, we ask one question for each order in the utterance, and the answer is the order's canonical paraphrase. We also include the previous answers when asking about the next order, to prevent the QA system from repeating the same orders. For single-turn QA, we ask only one question for all orders. We also include the canonical paraphrasing formulation for comparison. 

Thus, when canonical representations exist, single-turn QA and paraphrasing are  similar; the main difference is that our formulation always includes a context. However, we reiterate that canonicalization is often not viable in TOP. 

In summary, the general principle guiding the design of multi-turn interactions is that we first ask questions about intents, then we ask questions about their slots and slot values, and then repeat the process if we detect a nested intent in a slot. As for single-turn QA, we only ask one question and ensure that the answer encodes the entire tree.
When canonical grammars exist in a given domain, they can be used to train single-turn QA systems in a straightforward way.  

We experimentally evaluate the two QA variants and show how one may be preferred over the other under different settings.
\begin{figure}[t!]
  \begin{center}
    \begin{tabular}{c}
    \resizebox{.5\textwidth}{!}{
         \begin{forest}for tree={font=\huge}
          [ORDER [PIZZAORDER [NUMBER [\temv{two}] ] [SIZE [\temv{large}] ] [STYLE [\temv{everything}] ] ] [PIZZAORDER [NUMBER [\temv{two}] ] [SIZE [\temv{large}] ] [TOPPING [\temv{mushrooms}] ] [COMPLEX\textunderscore TOPPING [QUANTITY [\temv{extra}] ] [TOPPING [\temv{cheese}] ] ] ] [DRINKORDER [NUMBER [\temv{six}] ] [SIZE [\temv{large}] ] [DRINKTYPE [\temv{cokes}] ] ] ]
        \end{forest}
        }
    \end{tabular}
\end{center}
{\small \textbf{\textit{Canonical paraphrasing}}}:\\
\noindent\fbox{%
    \parbox{0.45\textwidth}{%
        \small
            \textbf{Grammar for drink orders:}\\
            I want {[number]} {[size]}/[volumn] [drinktype] [container]\\
            \textbf{Grammar for pizza orders:}\\
            I want [number] [size] pizza in the [style] style with [topping1; topping2; ...] with [complex\textunderscore topping] without [not\textunderscore topping1; not\textunderscore topping2; ...]\\
            \textbf{Input:} ``[utterance]''\\
            \textbf{Output:} I want two large pizza in the everything style; two large pizza with mushrooms with extra cheese; six large-sized cokes
        }
    }
{\small \textbf{\textit{Multi-turn QA}}}:\\
\noindent\fbox{%
    \parbox{0.46\textwidth}{%
        \small
            \textbf{Q}: A user said: ``[utterance]'' What order did the user place?\\
            {\red \textbf{$\mbox{\bf Q}\prime$:} A user said: ``[utterance]'' The user ordered \textbf{[MASK]}.}\\
            \textbf{A}: two large pizza in the everything style\\
            \textbf{Q}: A user said: ``[utterance]'' What order did the user place in addition to two large pizza in the everything style?\\
            {\red \textbf{$\mbox{\bf Q}\prime$:}: A user said: ``[utterance]'' The user ordered \textbf{[MASK]} in addition to two large pizza in the everything style.}\\
            \textbf{A}: two large pizza with mushrooms with extra cheese\\
            \textbf{Q}: A user said: ``[utterance]'' What order did the user place in addition to two large pizza in the everything style and two large pizza with mushrooms with extra cheese?\\
            {\red \textbf{$\mbox{\bf Q}\prime$:}: A user said: ``[utterance]'' The user ordered \textbf{[MASK]} in addition to two large pizza in the everything style and two large pizza with mushrooms with extra cheese.}\\
            \textbf{A}: six large-sized cokes\\
            \textbf{Q}: A user said: ``[utterance]'' What order did the user place in addition to two large pizza in the everything style and two large pizza with mushrooms with extra cheese and six large-sized cokes?\\
            {\red \textbf{$\mbox{\bf Q}\prime$:}: A user said: ``[utterance]'' The user ordered \textbf{[MASK]} in addition to two large pizza in the everything style and two large pizza with mushrooms with extra cheese and six large-sized cokes.}\\
            \textbf{A}: none
        }
    }
{\small \textbf{\textit{Single-turn QA}}}:\\
\noindent\fbox{%
    \parbox{0.46\textwidth}{%
        \small
            \textbf{Q}: A user said: ``...'' What orders did the user place?\\
            {\red \textbf{$\mbox{\bf Q}\prime$:}: A user said: ``...'' The user ordered \textbf{[MASK]}.}\\
            \textbf{A}: The user ordered two large pizza in the everything style; two large pizza with mushrooms with extra cheese; six large-sized cokes
        }
    }
    \caption{Parse tree, canonical paraphrasing formulation, multi-turn, and single-turn QA formulations for the utterance ``\textit{i would like two large pizzas with everything and two large pizzas with mushrooms and extra cheese and four large cokes}'' in the Pizza dataset. The {\red \textbf{$\mbox{\bf Q}^\prime$}} questions are used for training with the MSP (masked span prediction) objective.
    }
    \label{fig:pizza_qas}
\end{figure}

\subsection{Using MSP Objectives}
\citet{chada2021fewshotqa} have shown that fine-tuning pretrained seq2seq models to perform QA tasks with too few examples leads to much degraded performance, while training a QA model with a loss function directly aligned with the pretraining objective performs better. 
Inspired by this observation, we explore the following change to our QA formulation: instead of making QA a separate downstream task, we treat it as one of the pretraining tasks---masked span prediction, for which the models are trained to generate the entire masked span given one unique mask token~\cite{raffel2020exploring}.
Accordingly, instead of asking the model questions and having it generate arbitrary answers, we rephrase the question-answer pair as a declarative sentence where the answer is masked.
Thus, the model now has to denoise and recover the masked segment. 
We show an example of this approach in Fig.~\ref{fig:pizza_qas}.
The Q's are the original questions, whereas the primed Q's  in red are the declarative sentences with the answers masked out.
The answers to both Q's and primed Q's are the same, as mask tokens cover the exact answers.

\subsection{Converting Answers to Parse Trees}
\paragraph{Single-turn QA.}
In single-turn QA, each sentence of the answer concerns up to a fixed number of levels in the parse tree. Take the Topv2 instance in the bottom part of Fig.\ref{fig:qa_example_topv2} as a running example, where each sentence contains three levels: The first level is an intent node, the second level is its slots, and the third level is the value of each slot.
In the first sentence, which always corresponds to the root intent, we take what follows the phrase ``The user wants to''  as the intent, in the part before the first comma; the rest of the sentence is of the form ``where $S$ is $V$'', where $S$ is the slot and $V$ is the value. We note that one slot could have multiple values separated by semicolons, and for each value we create one slot node. In the case where an intent node doesn't have slot children, the sentence simply stops after the first part. 
To expand subsequent sentences into tree nodes, we make these sentences  start with ``The intent in [subutterance] is'', so we can traverse the parse tree to find the ``[subutterance]'' node and expand the tree from there.

\paragraph{Multi-turn QA.}
In the multi-turn model, reconstructing the parse tree is more straightforward. We categorize all questions into three groups: the first asking the intent for a (sub-)utterance, 
the second asking the slots that appear in an intent, and the third asking for the value given to a slot. We parse the questions for an utterance sequentially and identify the group to which each question belongs. When the question asks for the top-level intent, we build the root node. When it asks for the intent of a sub-utterance, we traverse the parse tree to look for the leaf node containing the exact text and add an intent child node (with the text as a node attribute) from there. For a question from the second group, we find the intent node whose attribute has the same sub-utterance and append the slots as its children. When the question asks for a slot value, we traverse the parse tree again, find which slot node they belong to, and add it as a child node. 

A final note for both single- and multi-turn QA: When we detect invalid entities generated by the QA models, we stop parsing and simply count such an instance as an incorrect output.

\section{Experiments}
We evaluate our method against two state-of-the-art seq2seq techniques, one generating linearized parse trees and the other generating canonical paraphrases.
We chose Topv2 and Pizza because they are the only two task-oriented datasets we are aware of with nested meaning representations that cannot be produced by slot-filling approaches. 
We investigate both multi- and single-turn QA. In the former, a question aims to recover one or several nodes in the parse tree; in single-turn QA, we only ask one question to reconstruct the entire tree. 
Additionally, we perform ablation analysis to evaluate the contribution of each component, as described in Sec.~\ref{sec:method}. We show that our method achieves superior performance, particularly on those TOPv2 utterances that have more nested semantics. 

\subsection{Datasets}
\paragraph{The Topv2 Dataset.}
The Topv2 dataset~\cite{chen2020low} is a collection of queries  produced by crowdsourced workers and intended for smart voice assistants.
Topv2 has compositional queries with hierarchical meaning representations and extends the original TOP dataset~\cite{gupta2018semantic} with six additional domains.
We present statistics for each domain in Table~\ref{tab:topv2_stats}.
The eight domains vary widely, including the number of samples (ranging from 13k to 31k), the number of slots (from 5 to 33), and the portion of flat utterances.
Accordingly, when we convert Topv2 instances to multi-turn QA instances, the average number of questions per instance varies a lot across the domains.
We test our QA approach on Topv2 under the full-data setting for all domains, and select four domains  to study in a few-shot setting, with only 10 samples per intent and slot (10SPIS).

\paragraph{The Pizza Dataset.}
Pizza is a new TOP dataset~\cite{pizzaDataset} consisting of complex utterances that order pizzas and drinks. Pizza consists of 2.4M training examples that are synthetically generated from a CFG, along with 348 dev examples and 1357 test examples generated and annotated by MTurk workers. 
Although the training set is large, we focus on few-shot settings with 30, 50, and 100 examples randomly drawn from the dev set.
The low-resource setting is indeed challenging---the dev set has only 107 unique slot values, whereas the test set has 180, requiring models to generalize well.

\subsection{Evaluation Metric}

We use a standard metric for evaluating TOP performance: unordered exact match accuracy (abbreviated as EM), which does a node-to-node comparison between the generated parse tree and the golden parse tree, modulo sibling order. EM doesn't take partially correct parses into account, so given a reference and a hypothesis, EM is either 0 or 1.

\subsection{Implementation Details}
We use the T5-large model~\cite{raffel2020exploring} as the backbone of our QA framework.
We choose a learning rate between 5e-6 and 5e-4, a batch size from \{32, 64, 96, 128\} for the full-data setting, and a batch size from \{8, 16, 24\} for the few-shot setting.
We search for the best set of hyperparameters with 16 random trials for each configuration. 
With full data, we train single-turn models for 10 epochs and multi-turn models for 30 epochs to select the best-performing checkpoint on validation. 
For few-shot learning, we train for 3000 steps and make a checkpoint every 100 steps. We used 8 Tesla V100 GPUs with 32 GB memory for all our training.

\subsection{Baselines}
We consider two baselines. The first applies a seq2seq model trained on logical forms (LFs) expressed in the ``TOP-decoupled format,'' which removes  from the parse tree all text that does not appear in a leaf slot~\cite{aghajanyan2020conversational}. For this baseline, we use the BART-Large model described by~\citet{lewis2020bart}, because this model is commonly used for TOP~\cite{rongali2020don,aghajanyan2020conversational,shrivastava2021span}, and we also use T5-large, because that is the backbone of our QA implementation. This allows us to eliminate any benefit that may be derived from the architecture itself. The second baseline method we consider generates canonical paraphrases of the original utterances~\cite{semanticmachines-2021-emnlp}, and we use T5-large to allow for a direct comparison. 

Note that for Topv2 we only compare our method against the first baseline, since that dataset doesn't have a canonical representation. Pizza does, so for that dataset we compare our method against both baselines (LFs in Top-decoupled notation and canonical paraphrases). 
Since we're not aware of any existing results on the baseline methods for individual Topv2 domains and for Pizza, we implemented both with HuggingFace~\cite{wolf-etal-2020-transformers} and performed hyperparameter tuning with the same computation budget given to our method.

\subsection{Main Results}
\begin{table}[t]
\centering
\resizebox{\columnwidth}{!}{
\begin{tabular}{lrrrrrr}
\toprule
\textbf{domain} & \multicolumn{1}{l}{\textbf{\#inst}} & \multicolumn{1}{l}{\textbf{\#int}} & \multicolumn{1}{l}{\textbf{\#slt}} & \multicolumn{1}{l}{\textbf{flat\%}} & \multicolumn{1}{l}{\textbf{depth}} & \multicolumn{1}{l}{\textbf{\#q/inst}} \\ \hline
alarm           & 30488                               & 8                                  & 9                                  & 84\%                                & 2.16                               & 3.28                                \\
event           & 13160                               & 11                                 & 17                                 & 80\%                                & 2.37                               & 6.55                                \\
message         & 14602                               & 12                                 & 27                                 & 84\%                                & 2.23                               & 3.97                                \\
music           & 17320                               & 15                                 & 9                                  & 100\%                               & 1.98                               & 2.81                                \\
navigation      & 30044                               & 17                                 & 33                                 & 57\%                                & 2.68                               & 6.18                                \\
reminder        & 26133                               & 19                                 & 32                                 & 79\%                                & 2.45                               & 7.67                                \\
timer           & 17392                               & 11                                 & 5                                  & 96\%                                & 2.00                               & 2.39                                \\
weather         & 31403                               & 7                                  & 11                                 & 100\%                               & 1.93                               & 3.16                               \\
\bottomrule
\end{tabular}
}
\caption{Domain statistics for Topv2. \#q/inst is the average number of questions per utterance in multi-turn QA. Navigation and reminder are two most nested domains. Music and weather have completely flat utterances.}
\label{tab:topv2_stats}
\end{table}

\begin{table*}[t!]
\centering
\resizebox{1.7\columnwidth}{!}{
\begin{tabular}{l|rrrrrrrrr}
\toprule
              & \multicolumn{1}{c}{\textbf{Alarm}} & \multicolumn{1}{c}{\textbf{Event}} & \multicolumn{1}{c}{\textbf{Message}} & \multicolumn{1}{c}{\textbf{Music}} & \multicolumn{1}{c}{\textbf{Navigation}} & \multicolumn{1}{c}{\textbf{Reminder}} & \multicolumn{1}{c}{\textbf{Timer}} & \multicolumn{1}{c}{\textbf{Weather}} & \multicolumn{1}{l}{\textbf{Average}} \\ \hline
BART LF  & 88.68                              & 85.66                              & 92.73                                & 80.35                              & 83.08                                   & 82.66                                 & 77.91                              & 89.81                                & 85.11                                \\
T5 LF    & 88.58                              & 85.12                              & 88.85                                & 82.60                               & 82.77                                   & 78.39                                 & 83.46                              & 89.95                                & 84.97                                \\ \hline
ST QA         & \textbf{90.38}                     & 88.58                              & 95.44                                & 82.83                              & 85.35                                   & 84.98                                 & 83.61                              & 91.94                                & 87.89                                \\
MT QA         & 90.05                              & 88.43                              & 96.81                                & 83.60                               & 82.07                                   & 87.72                                 & 81.84                              & 92.22                                & 87.84                                \\
MSP ST QA     & 90.27                              & \textbf{88.88}                     & 95.07                                & 82.95                              & \textbf{86.25}                          & 86.35                                 & \textbf{84.05}                     & 91.62                                & \textbf{88.18}                       \\
MSP MT QA     & 90.17                              & 88.43                              & \textbf{96.81}                       & \textbf{84.97}                     & 82.78                                   & \textbf{87.78}                        & 81.91                              & 92.34                                & 88.15                                \\ \hline
Relative Gain & 1.79\%                             & 3.76\%                             & 4.40\%                               & 2.87\%                             & 3.82\%                                  & 6.19\%                                & 0.71\%                             & 2.82\%                               & 3.29\%                    \\
\bottomrule
\end{tabular}
}
\caption{Topv2 results with full data. MSP refers to the model trained on masked span prediction. 
The relative gain for each domain is computed between the best QA variant and the best baseline method. 
The average is computed across eight domains. The best numbers are in bold.}
\label{tab:topv2_main_results}
\end{table*}

\begin{table}[t!]
\centering
\resizebox{\columnwidth}{!}{
\begin{tabular}{l|rrrrr}
\toprule
              & \multicolumn{1}{l}{\textbf{Alarm}} & \multicolumn{1}{l}{\textbf{Navigation}} & \multicolumn{1}{l}{\textbf{Reminder}} & \multicolumn{1}{l}{\textbf{Weather}} & \multicolumn{1}{l}{\textbf{Average}} \\ \hline
BART LF  & 45.82                              & 44.89                                   & 51.46                                 & 62.75                                & 51.23                                \\
T5 LF    & 59.34                              & 37.04                                   & 46.81                                 & 57.47                                & 50.17                                \\ \hline
ST QA         & 52.54                              & 28.39                                    & 48.58                                  & 52.25                                & 45.44                                \\ \vspace*{-.1cm}
MT QA         & 60.57                              & 55.56                                   & 69.72                                 & \textbf{81.89}                       & 66.94                                \\ \vspace*{-.15cm}
\hspace*{.3cm} w/o state    & {\small 50.22}                              & {\small 48.18}                                   & {\small 69.65}                                 & {\small 78.95}                                & {\small 61.75}                                \\ 
\hspace*{.3cm} w/o metadata      & {\small 52.19}                              & {\small 52.87}                                   & {\small 68.81}                                 & {\small 79.53}                                & {\small 63.35}                               \\
MSP ST QA     & \textbf{65.91}                     & 30.46                                   & 58.74                                 & 74.35                                & 57.37                                \\
MSP MT QA     & 62.13                              & \textbf{60.16}                          & \textbf{70.48}                        & 80.78                                & \textbf{68.39}                       \\ \hline
Relative Gain & 11.1\%                            & 34.0\%                                 & 37.0\%                               & 30.5\%                              & 28.1\%                             \\
\bottomrule
\end{tabular}
}
\caption{Topv2 results in the 10SPIS setting. For MT QA, we include results obtained by removing previous answers from the context (``w/o state'') and by removing domain metadata.}
\label{tab:topv2_lowshot_results}
\end{table}

\begin{table}[t!]
\centering
\resizebox{0.8\columnwidth}{!}{
\begin{tabular}{l|rrr}
\toprule
               & \multicolumn{1}{c}{\textbf{30}} & \multicolumn{1}{c}{\textbf{50}} & \multicolumn{1}{c}{\textbf{100}} \\ \specialrule{0.6pt}{1pt}{1pt}
BART LF   & 65.67                           & 72.00                           & 83.51                            \\
T5 LF     & 74.06                           & 77.82                           & 86.07                            \\
T5 Canonical   & 71.17                           & 77.67                           & 87.47                            \\ \hline
ST QA & 77.30                           & 81.58                           & 89.75                            \\
MT QA  & 66.96                  & 79.35                  & 85.98                   \\
MSP ST QA & 77.45                           & 83.14                           & 88.34                            \\
MSP MT QA  & \textbf{80.01}                  & \textbf{83.99}                  & \textbf{91.74}                   \\ \hline
Relative gain  & 8.03\%                          & 7.93\%                          & 4.96\%       \\
\bottomrule
\end{tabular}
}
\caption{Pizza results  in three few-shot settings (training with 30, 50, and 100 examples).}
\label{tab:pizza_results}
\end{table}

We first present  Topv2 results for both full-data and few-shot settings.
The baseline methods, labeled as BART/T5 LF, are trained on linearized Top-decoupled trees. We include four variants of the QA approach: ST QA and MT QA, as well as MSP ST QA and MSP MT QA (the same as ST and MT QA, except that the models are trained on the the masked span prediction objective).
We report the full-data EM scores in Table~\ref{tab:topv2_main_results} and the 10SPIS EM scores in Table~\ref{tab:topv2_lowshot_results}.
The relative gain is computed between the best QA approach and the best baseline approach in each domain.

In the full-data setting, all four QA variants outperform the baselines, with MSP ST QA having a slight overall edge,  exceeding the best baseline method by 3.07 absolute points.
We have the largest relative gain (6.19\%) in reminder.  In general, we see smaller improvements in the flatter domains: music, weather, timer, alarm are the four domains with the smallest semantic-tree depths, and relative gains for these domains are below 3\%.

We see that different QA approaches are close to each other when we have enough data; ST QA  performs only marginally better than MT QA. However, ST QA has shorter latency and may therefore be practically preferable. 

We choose four representative domains to perform low-resource experiments.  For 10SPIS, MSP MT QA is a clear winner over the other approaches, improving the best baseline method by 17.16 absolute points. ST QA has the worst performance. Our explanation is that ST QA requires the generation of long texts (longer than the Top-Decoupled LFs), and it is too challenging to learn a complex new task with only 10 instances per intent and slot. Therefore, in a few-shot setting, the benefit of having short answers is much clearer. Additionally, it is worth noting that having an objective that is well-aligned with the pretraining stage provides a significant benefit. It improves MT QA by nearly 5 absolute points on average, and it is a game changer for ST QA, in that with this one modification ST QA achieves competitive performance in three out of four domains.

We also test whether the inclusion of previous answers and metadata into the context helps in few-shot scenarios (the difference may be negligible with full data).  We remove answers to previous questions (``w/o state'') and metadata (``w/o metadata'') from MT QA and report the  scores obtained from both changes. The results suggest that both state and metadata make strong contributions to performance. Excluding prior answers has a more negative impact on average.

We note that our QA approach is very competitive against state-of-the-art results in the literature. For instance, the best reported result for 10SPIS reminder is 61.47~\cite{chen2020low}\footnote{We only mention reminder because \citet{chen2020low} don't report 10SPIS results for the other domains.}, which was achieved by first pretraining the model with six source domains. We improved that result by 9 absolute points while training only with the target domain's data. 

We next report results on Pizza, where we compare our approach against both LF-trained baselines (with LFs expressed in TOP-decoupled notation) and canonical paraphrases.
Table~\ref{tab:pizza_results} summarizes results for three low-resource settings. We obtain the greatest improvement with 30 training examples, for a relative gain exceeding 8\%. 
Consistently with the Topv2 results, as more training examples become available the gap between the QA approaches and the baselines shrinks. Even though T5 Canonical and ST QA are similar, the results show that including context information about an order is beneficial, especially when there are fewer training examples. MT QA has a slight edge over ST QA here, as asking more questions mitigates the burden of generating longer and more complex sequences.

\subsection{Additional Analysis: Depth vs. Accuracy}
\begin{table}[t!]
\centering
\resizebox{\columnwidth}{!}{
\begin{tabular}{l|rrrrrr}
\toprule
                            & \multicolumn{1}{l}{\textbf{D}} & \multicolumn{1}{l}{\textbf{L}} & \multicolumn{1}{l}{\textbf{\#inst}} & \multicolumn{1}{l}{\textbf{LF EM}} & \multicolumn{1}{l}{\textbf{QA EM}} & \multicolumn{1}{l}{\textbf{Rel. Gain}} \\ \hline
\multirow{5}{*}{Navigation} & 1 & 5.35                                 & 790                                 & 84.56                                 & 82.27                           & -2.71\%                                    \\
                            & 2 & 8.60                                 & 2689                                & 86.28                                 & 91.56                           & 6.12\%                                     \\
                            & 3 & 9.58                                 & 569                                 & 88.40                                  & 90.15                           & 1.98\%                                     \\
                            & 4 & 12.09                                 & 1905                                & 77.80                                  & 81.67                           & 4.97\%                                     \\
                            & 6 & 12.46                                 & 115                                 & 46.96                                 & 50.43                           & \textbf{7.39\%}                            \\ \hline
\multirow{3}{*}{Timer}      & 1 & 5.48                                 & 190                                 & 76.84                                 & 75.26                           & -2.06\%                                    \\
                            & 2 & 5.70                                 & 3746                                & 86.28                                 & 85.69                           & -0.68\%                                    \\
                            & 4 & 7.82                                 & 310                                 & 53.87                                 & 71.29                           & \textbf{32.34\%} \\        
\bottomrule
\end{tabular}
}
\caption{Topv2 full-data result analysis per depth level for navigation and timer. The average utterance length (L) and the number of test instances (\#inst) are shown for each depth (D). LF EM is from the best baseline method and QA EM is from the best QA method.}
\label{tab:depth_analysis}
\end{table}

We investigate how our approach performs on full-data Topv2 as a function of semantic depth.  We chose to perform the analysis on navigation and timer because navigation has the  greatest average tree depth, and we want to explain why our method gave the lowest improvements on timer.

We present the breakdown in Table~\ref{tab:depth_analysis}, where LF EM shows EM scores from the best baseline method generating TOP-decoupled trees, and QA EM shows EM scores from the best QA variant. For utterances with shallow semantics, generating TOP-decoupled trees outperforms QA; but as we move towards deeper semantic structures, the benefit of using a more naturalized representation becomes evident. Indeed, for both navigation and timer, QA performs best at the deepest level. Thus, for practical purposes, when the goal is to build a system that achieves high accuracy across all utterance groups, it is worth considering a combination of conventional LF generation for shallow utterances and a QA model for more complex utterances. We define the length of an utterance to be the number of words the utterance has. Table~\ref{tab:depth_analysis} lists the average utterance length for each depth, and shows that input size is a good proxy for semantic complexity (and could thus be used to quickly decide between the two approaches).

\section{Conclusion}
We have presented a reduction of compositional task-oriented parsing to abstractive QA, whereby parse tree nodes are recovered by posing queries to a QA model. 
We have also proposed to train QA models with the MSP (masked span prediction) objective, to better leverage the massive amount of linguistic knowledge gained during pretraining.
We experimentally evaluated single-turn QA and multi-turn QA on two public datasets, both in full-data and in few-shot settings, with and without MSP, and showed that they consistently outperform a number of powerful baseline techniques, including canonical paraphrasing, in both settings. The MSP variants perform best on average, with particularly dramatic improvements obtained in the few-shot setting.

\clearpage
\newpage

\bibliography{anthology}

\begin{thebibliography}{23}
\expandafter\ifx\csname natexlab\endcsname\relax\def\natexlab#1{#1}\fi

\bibitem[{Aghajanyan et~al.(2020)Aghajanyan, Maillard, Shrivastava, Diedrick,
  Haeger, Li, Mehdad, Stoyanov, Kumar, Lewis
  et~al.}]{aghajanyan2020conversational}
Armen Aghajanyan, Jean Maillard, Akshat Shrivastava, Keith Diedrick, Michael
  Haeger, Haoran Li, Yashar Mehdad, Veselin Stoyanov, Anuj Kumar, Mike Lewis,
  et~al. 2020.
\newblock Conversational semantic parsing.
\newblock In \emph{Proceedings of the 2020 Conference on Empirical Methods in
  Natural Language Processing (EMNLP)}, pages 5026--5035.

\bibitem[{Arkoudas et~al.(2021)Arkoudas, des Mesnards, Rubino, Swamy, Khanna,
  and Sun}]{pizzaDataset}
Konstantine Arkoudas, Nicolas~Guenon des Mesnards, Melanie Rubino, Sandesh
  Swamy, Saarthak Khanna, and Weiqi Sun. 2021.
\newblock \href
  {https://github.com/amazon-research/pizza-semantic-parsing-dataset} {Pizza: a
  task-oriented semantic parsing dataset}.

\bibitem[{Brown et~al.(2020)Brown, Mann, Ryder, Subbiah, Kaplan, Dhariwal,
  Neelakantan, Shyam, Sastry, Askell, Agarwal, Herbert-Voss, Krueger, Henighan,
  Child, Ramesh, Ziegler, Wu, Winter, Hesse, Chen, Sigler, Litwin, Gray, Chess,
  Clark, Berner, McCandlish, Radford, Sutskever, and
  Amodei}]{NEURIPS2020_1457c0d6}
Tom Brown, Benjamin Mann, Nick Ryder, Melanie Subbiah, Jared~D Kaplan, Prafulla
  Dhariwal, Arvind Neelakantan, Pranav Shyam, Girish Sastry, Amanda Askell,
  Sandhini Agarwal, Ariel Herbert-Voss, Gretchen Krueger, Tom Henighan, Rewon
  Child, Aditya Ramesh, Daniel Ziegler, Jeffrey Wu, Clemens Winter, Chris
  Hesse, Mark Chen, Eric Sigler, Mateusz Litwin, Scott Gray, Benjamin Chess,
  Jack Clark, Christopher Berner, Sam McCandlish, Alec Radford, Ilya Sutskever,
  and Dario Amodei. 2020.
\newblock \href
  {https://proceedings.neurips.cc/paper/2020/file/1457c0d6bfcb4967418bfb8ac142f64a-Paper.pdf}
  {Language models are few-shot learners}.
\newblock In \emph{Advances in Neural Information Processing Systems},
  volume~33, pages 1877--1901. Curran Associates, Inc.

\bibitem[{Chada and Natarajan(2021)}]{chada2021fewshotqa}
Rakesh Chada and Pradeep Natarajan. 2021.
\newblock Fewshotqa: A simple framework for few-shot learning of question
  answering tasks using pre-trained text-to-text models.
\newblock In \emph{Proceedings of the 2021 Conference on Empirical Methods in
  Natural Language Processing}, pages 6081--6090.

\bibitem[{Chen et~al.(2020)Chen, Ghoshal, Mehdad, Zettlemoyer, and
  Gupta}]{chen2020low}
Xilun Chen, Asish Ghoshal, Yashar Mehdad, Luke Zettlemoyer, and Sonal Gupta.
  2020.
\newblock Low-resource domain adaptation for compositional task-oriented
  semantic parsing.
\newblock In \emph{Proceedings of the 2020 Conference on Empirical Methods in
  Natural Language Processing (EMNLP)}, pages 5090--5100.

\bibitem[{Desai et~al.(2021)Desai, Shrivastava, Zotov, and Aly}]{desai2021low}
Shrey Desai, Akshat Shrivastava, Alexander Zotov, and Ahmed Aly. 2021.
\newblock Low-resource task-oriented semantic parsing via intrinsic modeling.
\newblock \emph{arXiv preprint arXiv:2104.07224}.

\bibitem[{Du et~al.(2021)Du, He, Li, Yu, Pasupat, and Zhang}]{du2021qa}
Xinya Du, Luheng He, Qi~Li, Dian Yu, Panupong Pasupat, and Yuan Zhang. 2021.
\newblock Qa-driven zero-shot slot filling with weak supervision pretraining.
\newblock In \emph{Proceedings of the 59th Annual Meeting of the Association
  for Computational Linguistics and the 11th International Joint Conference on
  Natural Language Processing (Volume 2: Short Papers)}, pages 654--664.

\bibitem[{Gao et~al.(2019)Gao, Sethi, Agarwal, Chung, and
  Hakkani-Tur}]{gao2019dialog}
Shuyang Gao, Abhishek Sethi, Sanchit Agarwal, Tagyoung Chung, and Dilek
  Hakkani-Tur. 2019.
\newblock Dialog state tracking: A neural reading comprehension approach.
\newblock In \emph{Proceedings of the 20th Annual SIGdial Meeting on Discourse
  and Dialogue}, pages 264--273.

\bibitem[{Guo et~al.(2018)Guo, Lu, Cai, Zhang, Yu, and Wang}]{guo2018long}
Jiaxian Guo, Sidi Lu, Han Cai, Weinan Zhang, Yong Yu, and Jun Wang. 2018.
\newblock Long text generation via adversarial training with leaked
  information.
\newblock In \emph{Proceedings of the AAAI Conference on Artificial
  Intelligence}, volume~32.

\bibitem[{Gupta et~al.(2018)Gupta, Shah, Mohit, Kumar, and
  Lewis}]{gupta2018semantic}
Sonal Gupta, Rushin Shah, Mrinal Mohit, Anuj Kumar, and Mike Lewis. 2018.
\newblock Semantic parsing for task oriented dialog using hierarchical
  representations.
\newblock In \emph{Proceedings of the 2018 Conference on Empirical Methods in
  Natural Language Processing}, pages 2787--2792.

\bibitem[{Lewis et~al.(2020)Lewis, Liu, Goyal, Ghazvininejad, Mohamed, Levy,
  Stoyanov, and Zettlemoyer}]{lewis2020bart}
Mike Lewis, Yinhan Liu, Naman Goyal, Marjan Ghazvininejad, Abdelrahman Mohamed,
  Omer Levy, Veselin Stoyanov, and Luke Zettlemoyer. 2020.
\newblock Bart: Denoising sequence-to-sequence pre-training for natural
  language generation, translation, and comprehension.
\newblock In \emph{Proceedings of the 58th Annual Meeting of the Association
  for Computational Linguistics}, pages 7871--7880.

\bibitem[{Li et~al.(2020)Li, Feng, Meng, Han, Wu, and Li}]{li2020unified}
Xiaoya Li, Jingrong Feng, Yuxian Meng, Qinghong Han, Fei Wu, and Jiwei Li.
  2020.
\newblock A unified mrc framework for named entity recognition.
\newblock In \emph{Proceedings of the 58th Annual Meeting of the Association
  for Computational Linguistics}, pages 5849--5859.

\bibitem[{Liu and Lane(2016)}]{liu16c_interspeech}
Bing Liu and Ian Lane. 2016.
\newblock \href {https://doi.org/10.21437/Interspeech.2016-1352}
  {{Attention-Based Recurrent Neural Network Models for Joint Intent Detection
  and Slot Filling}}.
\newblock In \emph{Proc. Interspeech 2016}, pages 685--689.

\bibitem[{Mansimov and Zhang(2021)}]{mansimov2021semantic}
Elman Mansimov and Yi~Zhang. 2021.
\newblock Semantic parsing in task-oriented dialog with recursive
  insertion-based encoder.
\newblock \emph{arXiv preprint arXiv:2109.04500}.

\bibitem[{McCann et~al.(2018)McCann, Keskar, Xiong, and
  Socher}]{mccann2018natural}
Bryan McCann, Nitish~Shirish Keskar, Caiming Xiong, and Richard Socher. 2018.
\newblock The natural language decathlon: Multitask learning as question
  answering.
\newblock \emph{arXiv preprint arXiv:1806.08730}.

\bibitem[{Namazifar et~al.(2021)Namazifar, Papangelis, Tur, and
  Hakkani-T{\"u}r}]{namazifar2021language}
Mahdi Namazifar, Alexandros Papangelis, Gokhan Tur, and Dilek Hakkani-T{\"u}r.
  2021.
\newblock Language model is all you need: Natural language understanding as
  question answering.
\newblock In \emph{ICASSP 2021-2021 IEEE International Conference on Acoustics,
  Speech and Signal Processing (ICASSP)}, pages 7803--7807. IEEE.

\bibitem[{Raffel et~al.(2020)Raffel, Shazeer, Roberts, Lee, Narang, Matena,
  Zhou, Li, and Liu}]{raffel2020exploring}
Colin Raffel, Noam Shazeer, Adam Roberts, Katherine Lee, Sharan Narang, Michael
  Matena, Yanqi Zhou, Wei Li, and Peter~J Liu. 2020.
\newblock Exploring the limits of transfer learning with a unified text-to-text
  transformer.
\newblock \emph{Journal of Machine Learning Research}, 21:1--67.

\bibitem[{Rongali et~al.(2022)Rongali, Arkoudas, Melanie, and
  Wael}]{RongaliLittleData2022}
Subendhu Rongali, Konstantine Arkoudas, Rubino Melanie, and Hamza Wael. 2022.
\newblock \href {https://arxiv.org/abs/2204.14243} {Training naturalized
  semantic parsers with very little data}.
\newblock In \emph{Proceedings of IJCAI 2022}.

\bibitem[{Rongali et~al.(2020)Rongali, Soldaini, Monti, and
  Hamza}]{rongali2020don}
Subendhu Rongali, Luca Soldaini, Emilio Monti, and Wael Hamza. 2020.
\newblock Don’t parse, generate! a sequence to sequence architecture for
  task-oriented semantic parsing.
\newblock In \emph{Proceedings of The Web Conference 2020}, pages 2962--2968.

\bibitem[{Shin et~al.(2021)Shin, Lin, Thomson, Chen, Roy, Platanios, Pauls,
  Klein, Eisner, and Durme}]{semanticmachines-2021-emnlp}
Richard Shin, Christopher~H. Lin, Sam Thomson, Charles Chen, Subhro Roy,
  Emmanouil~Antonios Platanios, Adam Pauls, Dan Klein, Jason Eisner, and
  Benjamin~Van Durme. 2021.
\newblock \href {http://cs.jhu.edu/~jason/papers/#semanticmachines-2021-emnlp}
  {Constrained language models yield few-shot semantic parsers}.
\newblock In \emph{Proceedings of the 2021 Conference on Empirical Methods in
  Natural Language Processing}, Punta Cana.

\bibitem[{Shrivastava et~al.(2021)Shrivastava, Chuang, Babu, Desai, Arora,
  Zotov, and Aly}]{shrivastava2021span}
Akshat Shrivastava, Pierce Chuang, Arun Babu, Shrey Desai, Abhinav Arora,
  Alexander Zotov, and Ahmed Aly. 2021.
\newblock Span pointer networks for non-autoregressive task-oriented semantic
  parsing.
\newblock \emph{arXiv preprint arXiv:2104.07275}.

\bibitem[{Wolf et~al.(2020)Wolf, Debut, Sanh, Chaumond, Delangue, Moi, Cistac,
  Rault, Louf, Funtowicz, Davison, Shleifer, von Platen, Ma, Jernite, Plu, Xu,
  Scao, Gugger, Drame, Lhoest, and Rush}]{wolf-etal-2020-transformers}
Thomas Wolf, Lysandre Debut, Victor Sanh, Julien Chaumond, Clement Delangue,
  Anthony Moi, Pierric Cistac, Tim Rault, Rémi Louf, Morgan Funtowicz, Joe
  Davison, Sam Shleifer, Patrick von Platen, Clara Ma, Yacine Jernite, Julien
  Plu, Canwen Xu, Teven~Le Scao, Sylvain Gugger, Mariama Drame, Quentin Lhoest,
  and Alexander~M. Rush. 2020.
\newblock \href {https://www.aclweb.org/anthology/2020.emnlp-demos.6}
  {Transformers: State-of-the-art natural language processing}.
\newblock In \emph{Proceedings of the 2020 Conference on Empirical Methods in
  Natural Language Processing: System Demonstrations}, pages 38--45, Online.
  Association for Computational Linguistics.

\bibitem[{Zhou et~al.(2021)Zhou, Naseem, Astudillo, and Florian}]{zhou2021amr}
Jiawei Zhou, Tahira Naseem, Ram{\'o}n~Fernandez Astudillo, and Radu Florian.
  2021.
\newblock Amr parsing with action-pointer transformer.
\newblock In \emph{Proceedings of the 2021 Conference of the North American
  Chapter of the Association for Computational Linguistics: Human Language
  Technologies}, pages 5585--5598.

\end{thebibliography}
\bibliographystyle{acl_natbib}


\end{document}